\documentclass{svproc}
\usepackage{url}
\usepackage[misc]{ifsym}
\usepackage{microtype}
\usepackage{graphicx}
\usepackage{subfigure}
\usepackage{booktabs} 
\usepackage{verbatim}
\usepackage{amsmath}
\usepackage{amssymb}
\usepackage{mathrsfs}
\usepackage{hyperref}
\usepackage{color}
\usepackage{algorithm}
\usepackage{algorithmicx}
\usepackage{algpseudocode}

\makeatletter
\def\thanks#1{\protected@xdef\@thanks{\@thanks
        \protect\footnotetext{#1}}}
\makeatother
\begin{document}
\mainmatter              
\title{Principal Gradient Direction and Confidence Reservoir Sampling for Continual Learning}
%
%
\author{Zhiyi Chen\inst{1} \and Tong Lin\textsuperscript{(\Letter)}\inst{2}}
%
%
%
\institute{Georgia Institute of Technology, USA\\
\email{zchen798@gatech.edu}
\and
The Key Laboratory of Machine Perception (MOE), School of EECS, Peking University; Peng Cheng Laboratory, Shenzhen, China\\
\email{lintong@pku.edu.cn}\\
}
\thanks{This work was supported by NSFC Tianyuan Fund for Mathematics (No. 12026606), and National Key R\&D Program of China (No. 2018AAA0100300).}
\maketitle              
\begin{abstract}
Task-free online continual learning aims to alleviate catastrophic forgetting of the learner on a non-iid data stream. Experience Replay (ER) is a SOTA continual learning method, which is broadly used as the backbone algorithm for other replay-based methods. However, the training strategy of ER is too simple to take full advantage of replayed examples and its reservoir sampling strategy is also suboptimal. In this work, we propose a general proximal gradient framework so that ER can be viewed as a special case. We further propose two improvements accordingly: \textbf{Principal Gradient Direction} (PGD) and \textbf{Confidence Reservoir Sampling} (CRS). In Principal Gradient Direction, we optimize a target gradient that not only represents the major contribution of past gradients, but also retains the new knowledge of the current gradient. We then present Confidence Reservoir Sampling for maintaining a more informative memory buffer based on a margin-based metric that measures the value of stored examples. Experiments substantiate the effectiveness of both our improvements and our new algorithm consistently boosts the performance of MIR-replay, a SOTA ER-based method: our algorithm increases the average accuracy up to 7.9\% and reduces forgetting up to 15.4\% on four datasets.
\end{abstract}
\section{Introduction}
Primates and humans can continually learn new skills and accumulate knowledge throughout their lifetime \cite{Fagot2006}. However, in machine learning, the agents hardly have a steady good performance when they learn a data stream. \emph{Catastrophic forgetting} \cite{McCloskey1989} is a common challenge when training a single neural network model on consecutive tasks: the model may perform well over the first task but suffers a serious accuracy decay along with the training process on the next tasks. \emph{Continual learning} \cite{Ring1994}, also known as \emph{lifelong learning} \cite{Thrun1994}, is a special field in machine learning that focuses on avoiding or alleviating catastrophic forgetting. 

The primary setting of continual learning (CL) is the task-incremental setting \cite{DBLP:journals/corr/abs-1904-07734}, which assumes the stream of data can be clearly divided into sequential tasks and learnt offline. However, task-free online  has received increasing attention recently, which is more practical: not only each sample can be merely observed once (single pass setting) but also the data stream is non-iid without any task information to assist the process of continual learning.

There are three major families of architecture in CL: expansion-based methods, regularization-based methods and replay-based methods. In this paper, we focus on the last one, which store the previous raw data and replay some of them when learning current data to alleviate forgetting. Experience Replay (ER) \cite{DBLP:journals/corr/abs-1902-10486} is one of the most representative methods, and has been proven as a strong baseline. Because of its superior performance, ER becomes the backbone algorithm for many recent replay-based methods, such as ER-MIR \cite{DBLP:journals/corr/abs-1908-04742}, GSS \cite{DBLP:conf/nips/AljundiLGB19}, etc. 

However, there is still room for improvement: on the one hand, the training strategy of ER is too simple to make full use of examples. On the other hand, reservoir sampling, which is a commonly used memory update strategy, can only ensure the equilibrium of previous samples but not good enough to maintain a more informative memory buffer. Our paper aims to tackle these defects and produces a stronger backbone algorithm for other continual learning methods based on ER.

In this paper, we firstly present a new algorithm for the training strategy called Principal Gradient Direction (PGD), which attempts to optimize a new gradient that not only represents the past data better but also retains the new knowledge of the current example. Secondly, we define a margin-based metric to measure the value of stored data and propose Confidence Reservoir Sampling (CRS), which helps to maintain a more informative memory buffer. 

Under the online CL setting, our experimental results show that both of our two approaches improve ER and also boost the performance of other ER-based CL methods, such as MIR \cite{DBLP:journals/corr/abs-1908-04742}, which achieve the best accuracy and forgetting measure among all the replay-based methods.

\section{Methods}
In this section, we will first discuss the setup of task-free online continual learning and replay-based methods in Section 2.1, and then propose a proximal gradient framework to analyze the training strategy of ER from a new perspective in Section 2.2. Finally, we elaborate our two methods: Principal Gradient Direction and Confidence Reservoir Sampling in Section 2.3 \& 2.4.
\subsection{Setup}
In task-free online continual learning setting, there is a stream of non-iid data: $...,(x_t, y_t),...$, which doesn't contain any task information to identify the specific task that one example belongs to. The learner $f(.;\theta^t)$ can only observe $(x_t, y_t)$ at the $t^{th}$ training step due to the single pass constraint. 

For replay-based methods, a space-limited memory buffer $\mathcal{M}$ can be used to store some examples to help provide information of past data. The learner should try to maximize the overall performance of all data, i.e., the average accuracy, and minimize the forgetting of past knowledge. 

Many methods \cite{DBLP:journals/corr/abs-1908-04742} \cite{DBLP:conf/nips/AljundiLGB19} have addressed their improvements on the simple random selection used in ER, which is orthogonal to our improvements. In the following subsections, we will analyze the shortcomings of ER on training strategy and storage strategy and present our improvements in Section 2.3 \& 2.4 accordingly. 

\subsection{Proximal Gradient Framework}
In this subsection, we use \emph{Proximal operator} \cite{Parikh2014}, a well-studied numerical method in optimization, to build a proximal gradient framework, which is the foundation of our Principal Gradient Direction and also provides a new perspective to the training strategy of ER. 

The proximal operator of a function $f(\cdot)$ with a scalar parameter $\lambda$ ($>0$) is defined by
\begin{equation}\label{proximal operator}
\text{prox}_{{\lambda}f}(v):={\arg\min}_{x}f(x)+\frac{1}{2{\lambda}}\|x-v\|_{2}^{2},
\end{equation}

where $x\in \mathbb{R}^n, v\in \mathbb{R}^n$ are two $n$ dimensional vectors and $f: \mathbb{R}^n \to \mathbb{R}$ is a closed proper convex function. Proximal operators can be interpreted as modified gradient steps:
\begin{equation}\label{proximal operator1}
\text{prox}_{{\lambda}f}(v)=v-\lambda\nabla M_{\lambda f}(v),
\end{equation}
where $M_{\lambda f}$ is a smoothed or regularized form of $f$ termed as Moreau envelop $M_{\lambda f}(v):=\inf_x f(x)+\frac{1}{2\lambda}\|x-v\|_2^2$.

As shown in \cite{DBLP:conf/nips/Lopez-PazR17}, continual learning can be formulated as a minimization problem that finds a new gradient close enough to the gradient of the new data and satisfies some constraints at the same time. In other words, the new gradient should still be beneficial to the current task and also takes the past tasks into consideration.

Based on this insight, we introduce the proximal operator into the setting of continual learning: 
\begin{equation}\label{proximal gradient}
\text{prox}_{{\lambda}f}(g)={\arg\min}_{w}f(w) + \frac{1}{2\lambda}\|g-w\|_{2}^{2},
\end{equation}

where $g$ is the gradient vector calculated on the new data, and $w$ is the target gradient to update the network weights. $f(\cdot)$ is the convex function we need to design which characterizes the relation between the target gradient and gradients of past examples selected from the memory. 

The training strategy of ER is simple: the learner randomly samples a small batch of past data from memory and directly uses the sampled data $B_t$ as well as the new input data $(x_t, y_t)$ to co-train the network. From the perspective of proximal gradient framework, the constraint function $f(\cdot)$ of ER is the inner product of the target gradient and the average gradient of selected past data without $\lambda$:
\begin{equation}\label{A-PGD-sum}
\min_{w}{\frac{1}{2}\|g-w\|_{2}^{2} - \langle g_{ref}, w \rangle},
\end{equation}

where $g_{ref}$ is the reference gradient of $B_t$. The Equation (4) has an analytic solution as follows, which is the actual training strategy of ER:
\begin{equation}\label{solution of A-PGD-sum}
w^* = g + g_{ref}.
\end{equation}

However, this strategy ignores the difference of sampled examples and it also regards new data and past data equally weighted, which is suboptimal.

\subsection{Principal Gradient Direction}
A more reasonable idea of utilizing the new data and selected examples is to find a target gradient that not only represents the overall contribution of the sampled past examples, but also maintains the knowledge of new data. Such a gradient can be found in the neighbor of $g$, which should also follow the principal direction of all past gradients. In this way, the new gradient will not violate the past knowledge for the reason that principal direction ensures a gradient descent towards a overall decrease on losses of past examples. In addition, the gradient also promotes the memorization of new data because it is a near neighbour of $g$.

To find the principal direction, we attempt to minimize the sum of solid angles between the new gradient vector and the past gradients, i.e., maximize the sum of cosine value. Besides, the length of a gradient should also be taken into consideration, because the ``short'' gradient vector means that current model $f(.;\theta^t)$ can learn it well and hence is less important than a ``long'' gradient. So we apply $sigmoid$ function on length of the gradient as weight. We can also set a small threshold $\epsilon$ for the length of gradient: $\left\|g_i\right\|_{\epsilon} = max(\epsilon, \left\|g_i\right\|)$ to further decrease the impact of the short one. 

Under the proximal gradient framework, we formulate a optimization problem as follows:
\begin{eqnarray}
\min_{w}  -\sum_{i=0}^{K}\frac{\langle w, g_i\rangle}{\left\|w\right\|\left\|g_i\right\|_{\epsilon}}sigmoid(\left\|g_i\right\|) + \frac{1}{2\alpha} \|w - g\|^2_2, 
\label{eq:origin}
\end{eqnarray}

where $w$ is the target gradient, $g$ is the gradient of the new input, $g_i$ is the gradient of the sampled past example, $\alpha$ is a hyperparameter to balance the two parts and $K$ is the size of sampled batch $B_n$. 

To solve this optimization problem, we choose \emph{Proximal Gradient Method} \cite{Parikh2014} to get an iterative solution of the proximal problem. Considering a general optimization problem:
\begin{eqnarray}
\mathrm{minimize} \quad f(x) + h(x),
\label{pgm}
\end{eqnarray}

where $f: \mathbb{R}^n \to \mathbb{R}$ and $h: \mathbb{R}^n \to \mathbb{R}\cup \{+\infty\}$ are two closed proper convex functions and $f$ is differentiable. The \emph{Proximal Gradient Method} is formulated as follows:
\begin{eqnarray}
x^{(k+1)} = prox_{\beta h}(x^{(k)}-\beta \nabla f(x^{(k)})).
\label{eq:pgm}
\end{eqnarray}

As for our problem, we regard the target gradient $w$ as the optimization variable, the principal direction term in (6) as function $f$ and the distance constraint term as function $h$. 

After substituting the variables and expanding the formulation of (8), we get the standard form of proximal gradient method for our optimization problem:
\begin{align}
w^{(k+1)} = &{\arg\min}_{w}\frac{1}{2{\beta}}\|w-(w^{(k)}-\beta \nabla f(w^{(k)}))\|_{2}^{2} \notag \\ 
&+\frac{1}{2\alpha} \|w-g\|_{2}^{2}.
\label{eq:pgm}
\end{align}

To find the solution, we need to set the derivative of (9) to zero. Note that we can ignore the constant term, e.g. $g^Tg$, so we can get:
\begin{align}
w^{(k+1)} = \frac{\alpha (w^{(k)}-\beta \nabla f(w^{(k)}))+\beta g}{\alpha+\beta}.
\end{align}

For the gradient $\nabla f(w^{(k)})$, with the rule of derivation for fraction, the solution is:
\begin{eqnarray}
\nabla f(w^{(k)}) = -\sum_{i=0}^{K}\left(\frac{g_i}{\left\|w^{(k)}\right\|\left\|g_i\right\|_{\epsilon}}sigmoid(\left\|g_i\right\|)\nonumber \right.\\
\left. -\frac{\langle w^{(k)}, g_i\rangle w^{(k)}}{\left\|w^{(k)}\right\|_2^3\left\|g_i\right\|_{\epsilon}}sigmoid(\left\|g_i\right\|)\right).
\end{eqnarray}

Here we choose the gradient of new input data $g$ as $w^{(0)}$ for the reason that the new gradient should be a neighbor of $g$. From empirical observation, we find that just one step optimization is good enough, so an approximate solution is:
\begin{eqnarray}
w^{(1)} = g - \frac{\alpha \beta}{\alpha + \beta}\nabla f(g).
\end{eqnarray}

We replace the fraction $\alpha \beta/ (\alpha + \beta)$ in (12) with a single hyperparameter $\lambda$ in experiment, which makes it look like one step gradient descent from $g$ on our principal direction function $f$. In practice, we can choose to group the examples averagely to decrease the number of backward propagation to obtain an appropriate computational complexity.

\subsection{Confidence Reservoir Sampling}
In this subsection, we focus on the storage strategy about how to update the memory with the new example $(x_t, y_t)$.

ER and many other replay-based methods apply reservoir sampling strategy (Algorithm \ref{reservoir sampling}) \cite{DBLP:journals/toms/Vitter85}, where mem\_sz is the total memory size of $\mathcal{M}$ and $t$ is the order number of input $(x_t, y_t)$. 

\begin{algorithm}[t]
\caption{\textbf{Reservoir sampling} }\label{reservoir sampling}
\begin{algorithmic}
    \State {\bfseries Procedure:} $\mathcal{M}$, mem\_sz, $t$, $(x_t,y_t)$
    \If {$|\mathcal{M}|$ $\leq$ mem\_sz}
    \State {$\mathcal{M}$.append(($x_t$,$y_t$))}
    \Else
    \State {$i$ = randint(0, $t$)}
    \If {$i$ $\leq$ mem\_sz}
    \State {\textcolor{blue}{$\mathcal{M}[i]\gets(x_t,y_t)$}}
    \EndIf
    \EndIf
\end{algorithmic}
\end{algorithm}

Though this strategy can ensure the equilibrium for memory buffer, the random replacement (the blue row in Algorithm \ref{reservoir sampling}) still has a room for improvement considering the limited memory space. We aspire to maintain a more informative memory buffer by replacing the less useful examples, which can improve continual learning no matter which subset is selected to consolidate the past knowledge. 

Just like the exploration and exploitation dilemma in reinforcement learning, the same situation also exists in online continual learning: exploration is replacing the old data with the new one to explore the new knowledge, while exploitation is keeping the old data intact. Actually, only when an example is selected, it is really exploited by the learner. 

Inspired by the idea of Upper-Confidence Bound (UCB) algorithm, which balances the uncertainty and reward of a certain action to choose one from the action set, we use a similar strategy to calculate a score for each example in memory buffer and choose the appropriate one to be replaced. 

The exploitation rate, denoted as $EX$, is the first part of the metric, which is calculated by a division from the times $n$ that the example is selected into $B_t$ and the age of the example $a$: $EX=n/a$. We intend to replace the highly exploited one, which is more likely to be overfitted by the learner. 

Then we define \emph{margin} \cite{Koltchinskii2002} based on the prediction probability from the forward propagation: the output prediction $p(x;\theta)$ on an example $(x,y)$ is computed through a softmax activation function, and we formulate margin, denoted as $m$, as:
\begin{equation}
m := p_{y}(x;\theta) - \max_{y'\neq y} p_{y'}(x;\theta).
\end{equation} 

When the model makes a correct prediction, the margin of the certain input is positive, otherwise, we get a negative margin. Margin value indicates the confidence of the prediction: larger the margin is in magnitude, more confidence we have in the prediction. 

At the $t^{th}$ training step, we can first get $m_t$ of $(x_t, y_t)$ from model $f(.;\theta^t)$ and then $m_{t+1}$ from the new model $f(.;\theta^{t+1})$ that executes one step gradient descent. Then we define margin increment: $MI = m_{t+1} - m_t$, which measures the importance of a certain example at one training step. If margin increment is large, it means that this training step has learnt the example very well, in other words, the example is simple and less informative for the model.

So we can calculate our metric, denoted as $\mathcal{S}$, for all the examples in memory buffer:
\begin{equation}
\mathcal{S} := EX + c\cdot MI,
\end{equation} 

where $EX$ is the exploitation rate, $MI$ is the margin increment and $c$ is a weight hyperparameter. For a high score, the example is either over-exploited or less informative, which is more appropriate to be replaced.

\begin{algorithm}[t]
\caption{\textbf{Confidence Reservoir sampling} }\label{confidence reservoir sampling}
\begin{algorithmic}
    \State {\bfseries Procedure:} $\mathcal{M}$, mem\_sz, $t$, $(x_t,y_t)$
    \If {$|\mathcal{M}|$ $\leq$ mem\_sz}
        \State {$\mathcal{M}$.append(($x_t$,$y_t$))}
    \Else
        \State {$i$ = randint(0, $t$)}
        \If {$i$ $\leq$ mem\_sz}
            \If {Using strategy $s_1$}
                \State {\textcolor{blue}{$j \gets \max(\mathcal{S(\mathcal{M})})$}}
            \ElsIf{Using strategy $s_2$} 
                \State {\textcolor{blue}{$j \sim P(j)= \mathcal{S}_j/\sum_k\mathcal{S}_k$}}
            \EndIf
            \State {\textcolor{blue}{$\mathcal{M}[j]\gets(x_t,y_t)$}}
        \EndIf
    \EndIf
\end{algorithmic}
\end{algorithm}

We have two strategies to replace examples based on $\mathcal{S}$: \\$s_{1}$ directly chooses the biggest score, and $s_2$ replaces each example with a probability $P(i)=\mathcal{S}_i/\sum_j\mathcal{S}_j$, which applies to different datasets.

So far, we complete the definition of our margin-based metric and implement it on reservoir sampling as Confidence Reservoir Sampling (Algorithm \ref{confidence reservoir sampling}). In this way, Confidence Reservoir Sampling not only satisfies the requirment of equal storage, but also maintains a more informative memory buffer. Note that our margin-based metric can also be extended to other storage strategy.

\section{Experiments}
In this section, we report the details of experiments and the performance of our two improvements. We apply PGD and CRS on ER and conduct ablation study. We also use the renewed backbone algorithm over MIR-replay \cite{DBLP:journals/corr/abs-1908-04742} to demonstrate the effectiveness of our approaches.

\subsection{Datasets and Architectures}
We consider four commonly used datasets:  \\
(1) \textbf{MNIST Split} is derived from MNIST, the famous dataset on handwritten digits, which directly splits 10 classes of MNIST into 5 non-overlapping different tasks. \\
(2) \textbf{MNIST Permutations} is also derived from MNIST, which randomly generates different pattern of pixel permutation for each task to exchange the position of the original images of MNIST. For both MNIST Split and MNIST Permutations, we use the similar benchmark setting as \cite{DBLP:conf/nips/Lopez-PazR17} that each task consists of 1000 examples.\\
(3) \textbf{CIFAR10 Split} is derived from CIFAR10, which averagely divides the whole classes in CIFAR10 into 5 tasks, where each task has 9750 samples and 250 retained for validation just as \cite{DBLP:journals/corr/abs-1908-04742}.\\
(4) \textbf{MiniImageNet Split} is derived from miniImageNet, a subset of ImageNet with 100 classes and 600 images per class, which averagely divides the whole classes into 20 tasks.  

For MNIST-S and MNIST-P, all baselines use fully-connected neural networks with two hidden layers of $100$ ReLU units. A smaller version of ResNet18 \cite{DBLP:conf/cvpr/HeZRS16} is used for CIFAR10-S and MINI-S, which has three times less feature maps for each layer than the original ResNet18.

\begin{table}[t]
\caption{Average accuracy (\%) of ablation Study ($\uparrow$)} 
\label{ablation}
\vskip 0.15in
\begin{center}
\begin{small}
\begin{sc}
\setlength{\tabcolsep}{1mm}{
\begin{tabular}{lcccc}
\toprule
Method & ER & ER-P & ER-C & ER-PC \\
\midrule
MNIST-S    & 79.8$\pm$3.2& 82.4$\pm$2.1 & 81.5$\pm$2.1 & \textbf{84.0$\pm$2.3} \\
MNIST-P & 79.1$\pm$0.7& 80.9$\pm$0.3& 79.9$\pm$0.5 & \textbf{81.7$\pm$0.6}\\
CIFAR10-S    & 30.7$\pm$2.0& 36.1$\pm$1.8& 38.5$\pm$1.1 & \textbf{40.0$\pm$2.1} \\
Mini-S    & 23.0$\pm$1.2& 25.5$\pm$0.6& 25.2$\pm$0.8 & \textbf{25.8$\pm$1.0}    \\
\bottomrule
\end{tabular}}
\end{sc}
\end{small}
\end{center}
\vskip -0.1in
\end{table}
\begin{table}[t]
\caption{Forgetting measure (\%) of ablation Study ($\downarrow$)}
\label{ablation}
\vskip 0.15in
\begin{center}
\begin{small}
\begin{sc}
\setlength{\tabcolsep}{1mm}{
\begin{tabular}{lrrrr}
\toprule
Method & ER & ER-P & ER-C & ER-PC \\
\midrule
MNIST-S    & 19.2$\pm$4.0& 13.2$\pm$3.1 & 17.3$\pm$3.2 &\textbf{9.6$\pm$1.9} \\
MNIST-P & 4.3$\pm$0.5& 2.6$\pm$0.5& 4.0$\pm$0.6 & \textbf{2.4$\pm$0.4}\\
CIFAR10-S    & 63.3$\pm$2.7& 56.6$\pm$3.7& \textbf{49.4$\pm$1.7} & 49.7$\pm$3.3 \\
Mini-S    & 32.1$\pm$2.0& 25.7$\pm$1.4& 28.5$\pm$1.1 & \textbf{25.7$\pm$1.1}    \\
\bottomrule
\end{tabular}}
\end{sc}
\end{small}
\end{center}
\vskip -0.1in
\end{table}

\subsection{Metrics}
We use \emph{Average Accuracy} and \emph{Forgetting Measure} \cite{DBLP:conf/iclr/ChaudhryRRE19} to evaluate the performances of the baselines over four datasets. For \emph{Average Accuracy}, the higher the number (indicated by $\uparrow$) the better is the model. For \emph{Forgetting Measure}, the lower the number (indicated by $\downarrow$) the better is the model. We run 10 times to get each result. 

\subsection{Ablation Study}
We conduct ablation study on four datasets by combining our two approaches with ER, and the resulting algorithms are as follows: basic ER (noted as ER), ER pluses PGD (noted as ER-P), ER pluses CRS (noted as ER-C) and ER pluses both PGD and CRS (noted as ER-PC). We store 50 examples per class and select 10 past examples for $B_t$ on MNIST-S, MNIST-P and CIFAR10-S while store 100 examples per class and select 20 examples on MINI-S. The results are showed in Table 1 \& 2.

\textbf{Effectiveness of PGD and CRS} From the results, we can observe that both PGD and CRS can improve the performance of ER on all four datasets: the two methods can boost the average accuracy up to 7.7\% and reduce the forgetting measure up to 13.9\%. On MNIST-S and MNIST-P, whose size are relatively small and network is simpler, PGD contributes more than CRS. The situation reverses on CIFAR10-S. MINI-S has the longest task sequence (20 tasks) and the biggest input size, where our two approaches have similar contribution in average accuracy. The comparative relations are same in forgetting measure.

\textbf{Joint improvement of PGD and CRS} The results also demonstrate that PGD and CRS can always jointly render a further improvement. On all four datasets, ER-PC is the best algorithm in terms of average accuracy which outperforms ER from 2.6\% to 9.3\%. ER-PC also achieves least forgetting on the first three datasets, which only performs slightly worse than ER-C on CIFAR10-S.  

\begin{figure}[t]
\centering
\begin{minipage}[t]{0.48\textwidth}
\centering
\includegraphics[width=5cm]{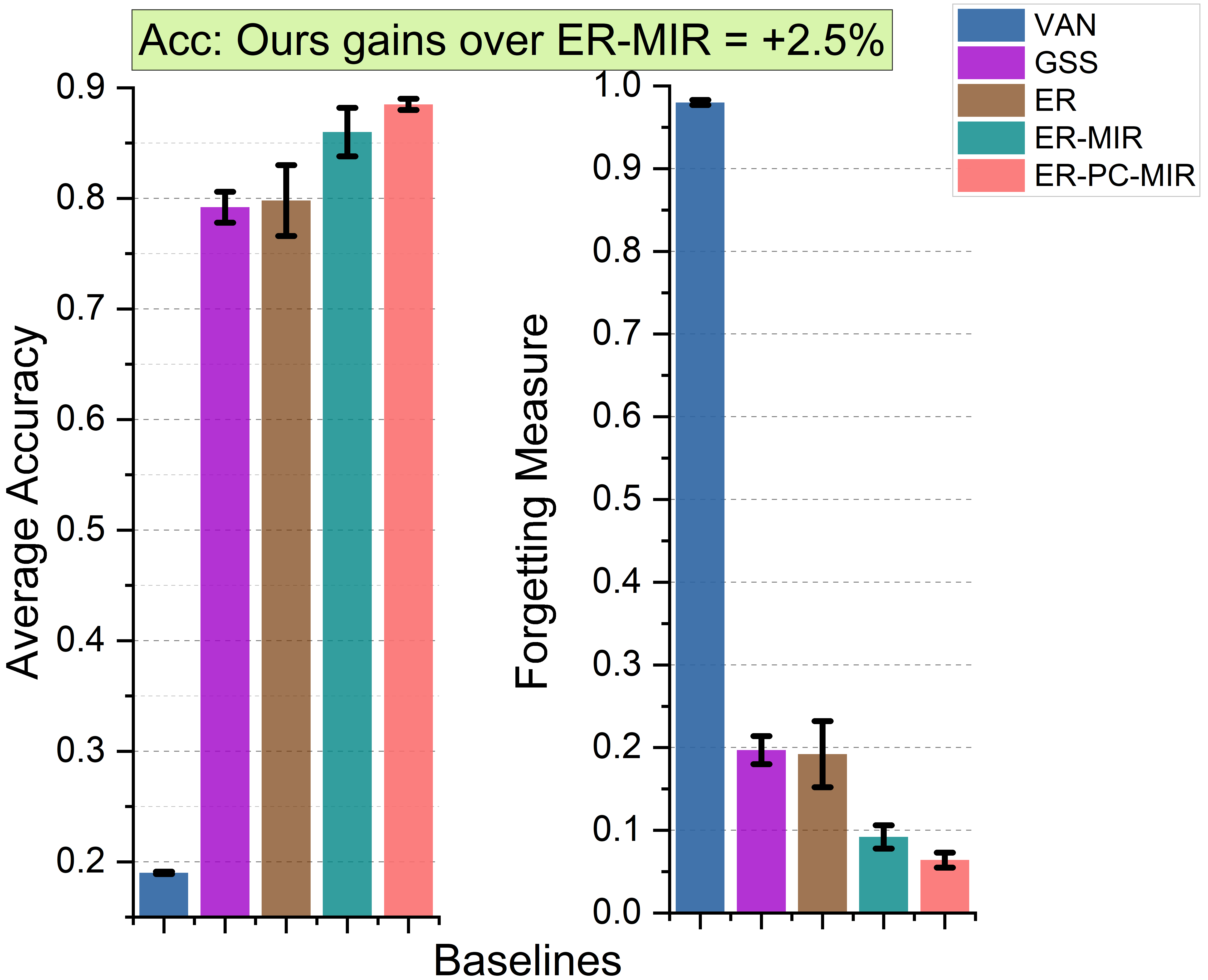}
\caption{Performances on MNIST-S}
\end{minipage}
\begin{minipage}[t]{0.48\textwidth}
\centering
\includegraphics[width=5cm]{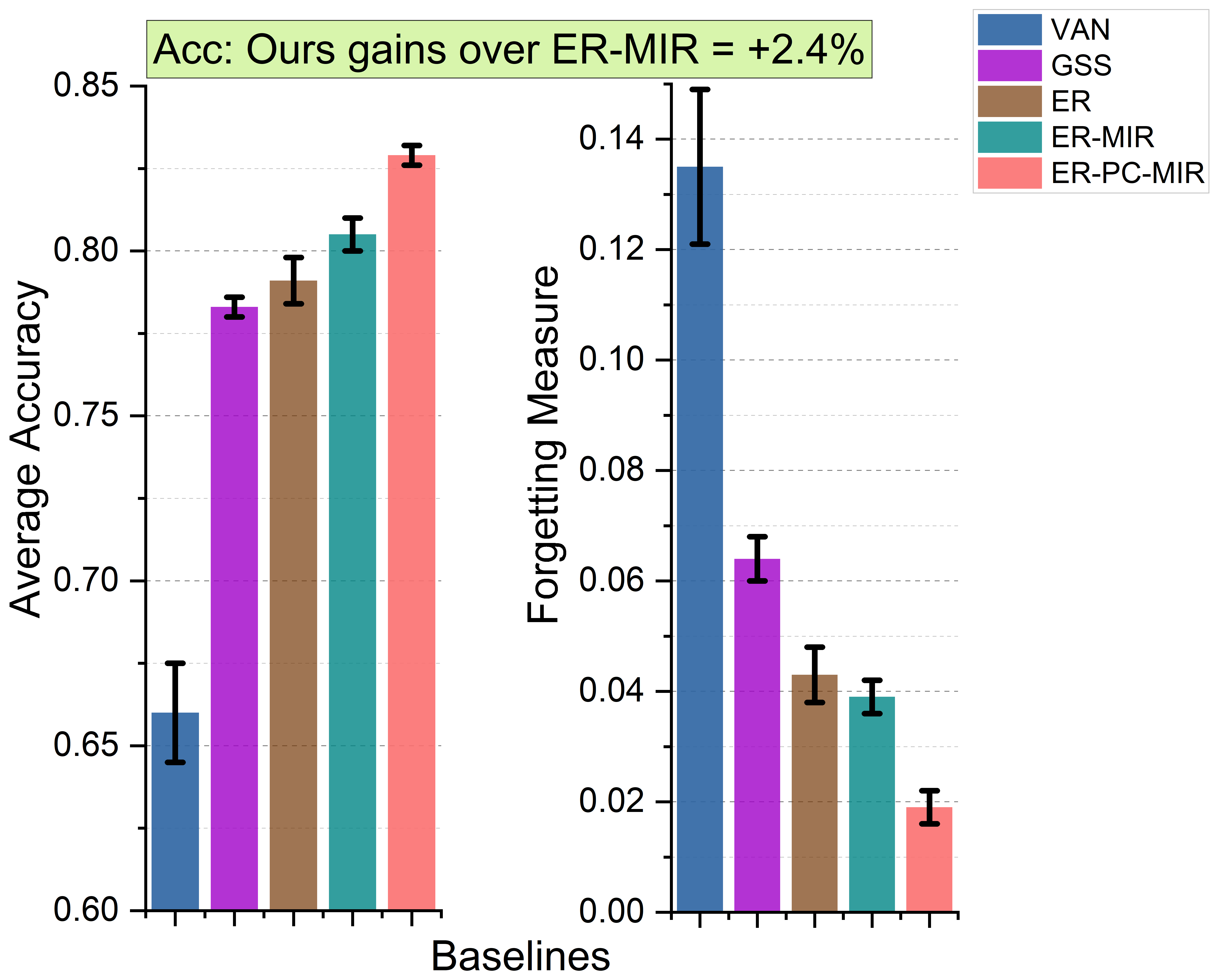}
\caption{Performances on MNIST-P}
\label{figure1}
\end{minipage}
\end{figure}

Our aim is to produce a stronger backbone algorithm for other ER-based methods, so we use ER-PC as a renewed backbone algorithm for the following comparison.

\subsection {Performance of ER-PC}
In this subsection, we will show the performance of ER-PC, where we use it as the new backbone algorithm by overlying MIR-replay \cite{DBLP:journals/corr/abs-1908-04742} on it, which is an example-selection strategy for replay and is SOTA replay-based method so far. We note the new method as \textbf{ER-PC-MIR}. 

\begin{figure}[t]
\centering
\begin{minipage}[t]{0.48\textwidth}
\centering
\includegraphics[width=5cm]{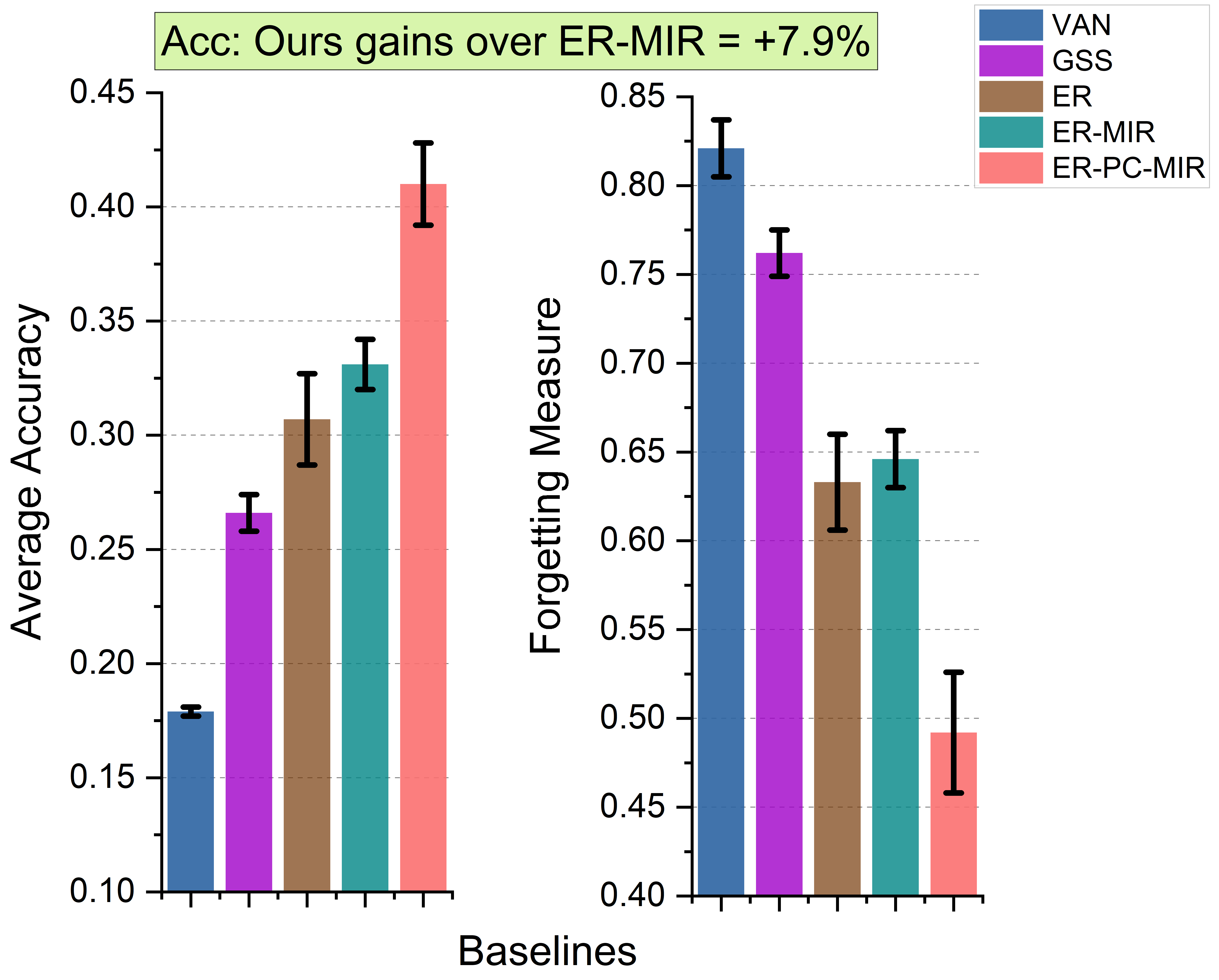}
\caption{Performances on CIFAR10-S}
\end{minipage}
\begin{minipage}[t]{0.48\textwidth}
\centering
\includegraphics[width=5cm]{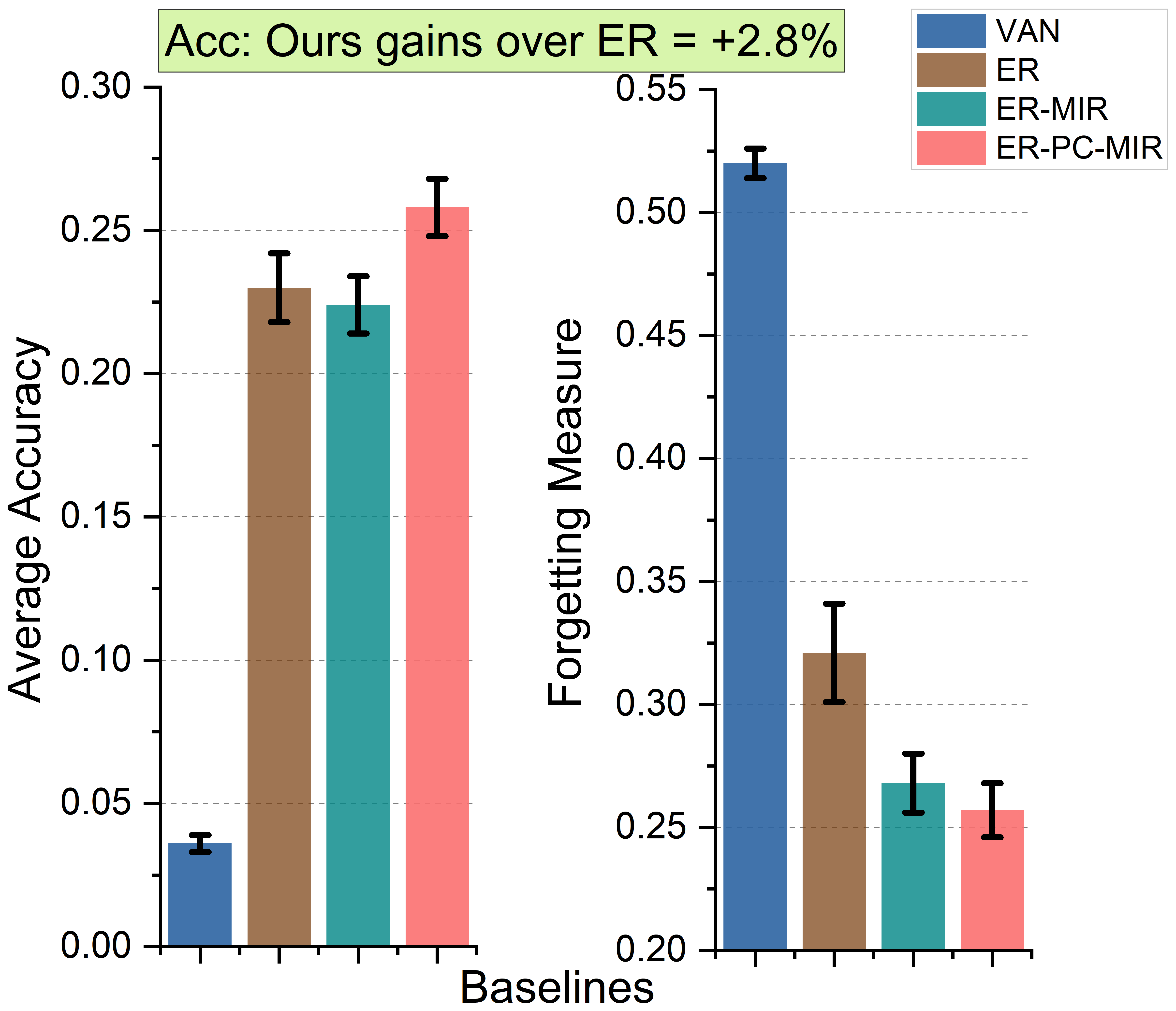}
\caption{Performances on MINI-S}
\label{figure1}
\end{minipage}
\end{figure}

\subsubsection{Basic comparison}
We take the following four baselines into comparison: 
\textbf{VAN} (a vanilla method that a single predictor for all the tasks without any continual learning strategy), 
\textbf{ER} \cite{DBLP:journals/corr/abs-1902-10486}, \textbf{ER-MIR} \cite{DBLP:journals/corr/abs-1908-04742} (the basic version of MIR-replay based on ER) and \textbf{GSS}  \cite{DBLP:conf/nips/AljundiLGB19}. 

For the reason that the training time of GSS on MINI-S is unacceptable, we don't take GSS into comparison on this dataset. We also don't take GEM \cite{DBLP:conf/nips/Lopez-PazR17} and A-GEM \cite{DBLP:conf/iclr/ChaudhryRRE19} into comparison because they all need the task information to update the memory and train the network, which violate the task-free online CL setting. Prior works show that ER and ER-MIR outperform GEM-like algorithms. The settings of memory size are same as our ablation study. The results are reported in Figure 1 - 4.

First, ER-PC-MIR achieves the best average accuracy on all four datasets. On MNIST-S, MNIST-P and CIFAR10-S, ER-PC-MIR achieves better average accuracy than ER-MIR, the best baseline on these datasets, with improvements up to 7.9\%. In MINI-S, our method is better than ER, the best baseline, with improvement 2.8\%.

Second, our method also forgets least knowledge among the baselines on all four datasets: ER-PC-MIR reduces forgetting than ER-MIR with improvements from 1.1\% to 15.4\%. On CIFAR10-S, ER is the best baseline in terms of forgetting measure, and ER-PC-MIR is better than it with 13.6\%.

The results show that our method ER-PC is a stronger backbone algorithm than vanilla ER: after combining with MIR-replay, ER-PC-MIR not only outperforms than ER-MIR, but also achieves the best performance among all other replay-based methods.

\subsubsection{Comparison in Different Memory Size}
As MNIST-P and CIFAR10-S are two representative datasets in domain-incremental and class-incremental datasets, we run ER-MIR and ER-PC-MIR on them in different memory size. We store 100, 50, 25 and 10 examples per class, which means that the total size of memory buffer is 1000, 500, 250, 100 on two datasets. We report the average accuracy and forgetting measure in Table 3 \& 4.
\begin{table}[t]
\caption{Average accuracy (\%) on MNIST-P and CIFAR10-S in different memory size ($\uparrow$)}
\label{ablation}
\vskip 0.15in
\begin{center}
\begin{small}
\begin{sc}
\setlength{\tabcolsep}{1mm}{
\begin{tabular}{lcccc}
\toprule
MNIST-P & 1000 & 500 & 250 & 100 \\
\midrule
ER-MIR & 82.7$\pm$0.4& 80.5$\pm$0.5& 77.5$\pm$0.9 & 73.6$\pm$1.0\\
ER-PC-MIR    & \textbf{84.4$\pm$0.4}& \textbf{82.9$\pm$0.3}& \textbf{79.6$\pm$0.6} & \textbf{76.1$\pm$0.4} \\ \midrule
CIFAR10-S & 1000 & 500 & 250 & 100 \\
\midrule
ER-MIR & 43.5$\pm$1.7& 33.1$\pm$1.1& 27.1$\pm$2.3 & 22.0$\pm$2.2\\
ER-PC-MIR    & \textbf{48.9$\pm$2.5}& \textbf{41.0$\pm$1.8}& \textbf{33.7$\pm$1.9} & \textbf{26.6$\pm$3.0} \\
\bottomrule
\end{tabular}}
\end{sc}
\end{small}
\end{center}
\vskip -0.1in
\end{table}

\begin{table}[t]
\caption{Forgetting measure (\%) on MNIST-P and CIFAR10-S in different memory size ($\downarrow$)}
\label{ablation}
\vskip 0.15in
\begin{center}
\begin{small}
\begin{sc}
\setlength{\tabcolsep}{1mm}{
\begin{tabular}{lcccc}
\toprule
MNIST-P & 1000 & 500 & 250 & 100 \\
\midrule
ER-MIR & 2.3$\pm$0.4& 3.9$\pm$0.3& 6.0$\pm$0.6 & 8.8$\pm$0.9\\
ER-PC-MIR    & \textbf{1.2$\pm$0.3}& \textbf{1.9$\pm$0.3}& \textbf{4.4$\pm$0.5} & \textbf{7.0$\pm$0.7} \\\midrule
CIFAR10-S & 1000 & 500 & 250 & 100 \\
\midrule
ER-MIR & 46.4$\pm$ 5.1& 64.6$\pm$1.6& 72.2$\pm$4.3 & 77.0$\pm$2.5\\
ER-PC-MIR    & \textbf{36.0$\pm$4.8}&\textbf{49.2$\pm$3.4}& \textbf{54.6$\pm$3.8} & \textbf{69.1$\pm$5.3} \\
\bottomrule
\end{tabular}}
\end{sc}
\end{small}
\end{center}
\vskip -0.1in
\end{table}
In all memory size, ER-PC-MIR consistently improves the performance of ER-MIR. ER-PC-MIR achieves more average accuracy than ER-MIR from 1.7\% to 2.5\% on MNIST-P. On CIFAR10-S, ER-PC-MIR gains over ER-MIR from 4.6\% to 7.9\% in average accuracy. The results show the reliability of our renewed backbone algorithm in different memory size.

\section{Conclusion}
In this paper, we firstly focus on the training strategy of CL and present a proximal gradient framework. Based on it, \textbf{Principal Gradient Direction} is proposed to take full advantage of replayed examples and new data. Then we pay attention to memory updating strategy: we define a new margin-based metric to measure the value of stored data and propose \textbf{Confidence Reservoir Sampling} based on it to maintain a more informative memory buffer. The experiments demonstrate that our two approaches are both beneficial and can jointly give a further improvement. After applied with PGD and CRS, the renewed backbone algorithm can boost the performance of MIR-replay and always achieves the best performance among other replay-based baselines on four datasets. On task-incremental and domain-incremental datasets, our method also consistently outperforms ER-MIR in different memory size. The experiments show that our method is a reliable and stronger backbone algorithm than vanilla ER.

%
%

\end{document}